\documentclass[letterpaper, 10 pt, conference]{ieeeconf}  % Comment this line out if you need a4paper

\IEEEoverridecommandlockouts                              % This command is only needed if 
\usepackage[T1]{fontenc}
\usepackage{blindtext}
\usepackage{graphicx}
\usepackage[table]{xcolor}

\usepackage{amsmath}
\usepackage{amssymb}
\usepackage{graphicx}
\usepackage{placeins}
\usepackage{diagbox} 
\usepackage[caption=false,font=footnotesize]{subfig}
\usepackage{gensymb}
\usepackage{textcomp}
\usepackage[pdfencoding=auto, hidelinks]{hyperref}
\usepackage{amsmath}
\usepackage{multirow}
\usepackage[hang,flushmargin]{footmisc}
\usepackage{microtype}
\usepackage{flushend}

\usepackage{amsmath}
\DeclareMathOperator*{\argmax}{arg\,max}
\DeclareMathOperator*{\argmin}{arg\,min}

\usepackage{mathtools}

\usepackage[colorinlistoftodos]{todonotes}

\usepackage[free-standing-units=true]{siunitx}

\newcommand{\eg}{\mbox{e.\,g.}\xspace}

\newcommand{\etal}{\emph{et~al.}\xspace}
   
\renewcommand{\[}{\begin{equation}}
\renewcommand{\]}{\end{equation}}

\renewcommand{\eqref}[1]{Eq.~(\ref{#1})}
\newcommand{\figref}[1]{Fig.~\ref{#1}}

\renewcommand{\baselinestretch}{0.99}

\DeclareMathAlphabet{\mathcal}{OMS}{cmsy}{m}{n}

\def\I{\ensuremath{\mathbb{I}}}

\renewcommand{\vec}[1]{\ensuremath{\mathbf{#1}}}
\newcommand{\mat}[1]{\ensuremath{\mathbf{#1}}}
\newcommand{\norm}[1]{\left\lVert#1\right\rVert}
\newcommand{\fd}[1]{\ensuremath{\dot{#1}}}

\newcommand{\calA}{{\cal A}}

\newcommand{\calK}{{\cal K}}

\newcommand{\calP}{{\cal P}}

\newcommand{\calX}{{\cal X}}
\newcommand{\calZ}{{\cal Z}}

\newcommand{\World}{\mathtt{W}}
\newcommand{\EndEff}{\mathtt{E}}
\newcommand{\Grasp}{\mathtt{G}}
\newcommand{\Articulation}{\mathtt{A}}

\newcommand{\R}{\mathbf{R}}

\newcommand{\T}{\mathbf{T}}
\newcommand{\Real}{\mathbb{R}}

\newcommand{\SE}{\mathrm{SE}}
\newcommand{\SEthree}{\SE(3)}

\newcommand{\State}{\boldsymbol{x}}

\newcommand{\linvel}{\mathbf{v}}
\newcommand{\rotvel}{\boldsymbol{\omega}}
\newcommand{\xihat}{\hat{\xi}}
\newcommand{\part}{\mathbf{T}}
\newcommand{\partA}{\mathbf{T}^{A}}
\newcommand{\partB}{\mathbf{T}^{B}}

\title{\LARGE \bf
Online Estimation of Articulated Objects with\\ Factor Graphs using Vision and Proprioceptive Sensing
}

\author{Russell Buchanan$^{1}$, Adrian R\"ofer$^{2}$, Jo\~{a}o Moura$^{1}$, Abhinav Valada$^{2}$, and Sethu Vijayakumar$^{1}$
\thanks{$^{1}$ School of Informatics, University of Edinburgh, UK}
\thanks{$^{2}$ Department of Computer Science, University of Freiburg, Germany}
\thanks{\noindent This work is supported by the Alan Turing Institute, EU H2020 Project Harmony (Grant No. 101017008), the Carl Zeiss Foundation ReScaLe project, and the BrainLinks-BrainTools center of the University of Freiburg.}
}

\begin{document}

\maketitle

\begin{abstract}
From dishwashers to cabinets, humans interact with articulated objects every day, and for a robot to assist in common manipulation tasks, it must learn a representation of articulation. 
Recent deep learning learning methods can provide powerful vision-based priors on the affordance of articulated objects from previous, possibly simulated, experiences. 
In contrast, many works estimate articulation by observing the object in motion, requiring the robot to already be interacting with the object. 
In this work, we propose to use the best of both worlds by introducing an online estimation method that merges vision-based affordance predictions from a neural network with interactive kinematic sensing in an analytical model. 
Our work has the benefit of using vision to predict an articulation model before touching the object, while also being able to update the model quickly from kinematic sensing during the interaction. 
In this paper, we implement a full system using shared autonomy for robotic opening of articulated objects, in particular objects in which the articulation is not apparent from vision alone. 
We implemented our system on a real robot and performed several autonomous closed-loop experiments in which the robot had to open a door with unknown joint while estimating the articulation online. 
Our system achieved an 80\% success rate for autonomous opening of unknown articulated objects.
\end{abstract}

\section{Introduction}
\begin{figure}[t]
	\centering
	\includegraphics[width=\columnwidth]{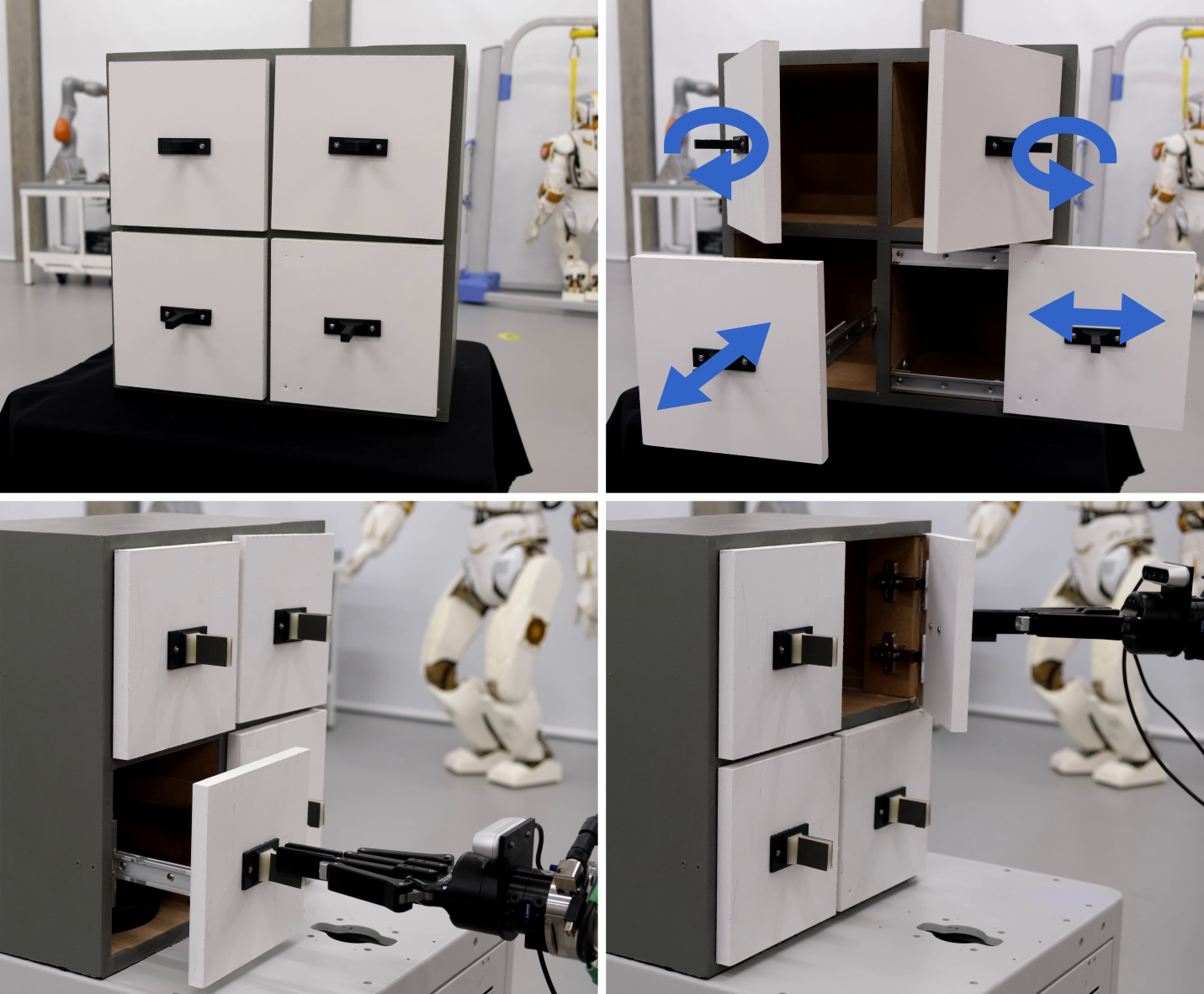}
	\caption{Top row: a cabinet with a set of visually identical doors. Their different articulations are only revealed once open. It would not be possible from visual inspection alone to predict how each door opens. Bottom row: the robot autonomously opens the cabinet while estimating articulation online.}
	\label{fig:motivation}
\end{figure}

Articulated objects are ubiquitous in everyday environments: dishwashers, microwaves, and cabinets are all objects that robots need to interact with to perform useful tasks like cooking or cleaning. 
To interact with these objects, a robot needs an understanding of the articulation -- either as an analytical model (\eg revolute or prismatic joint) or as an implicitly learned model through a neural network. Many recent works apply deep learning to predict affordance from vision; however, articulation prediction from vision alone may not be feasible.
For example, in \figref{fig:motivation}, each door appears identical until a person or robot interacts with them.
This is a problem for robotic systems which rely exclusively on vision for understanding articulation.
In our work, we propose a method for joint optimization of vision and proprioceptive sensing for online estimation of articulated objects. 

Some early work on manipulating articulated objects focused on parts-based estimation using only proprioception~\cite{Jain2010}.
Their method achieved impressive reliability in 2010, opening commonplace doors and drawers using only kinematic sensing and a grasp pose provided by a user. 
Vision has also been used, usually by tracking features or fiducial markers on the moving part and estimating articulation from the observed motion~\cite{Tresadern2005, Sturm2011}.
This can be combined with proprioceptive sensing to manipulate a wider variety of objects~\cite{MartinMartin2022}.
The limitation of these methods is that sensing (both vision and proprioceptive) only occurs while the articulated part is in motion.
The robot has no initial guess of how the object opens and so must rely on human direction or random guessing.

In response to this, the research focus has shifted recently to deep learning with only visual information used to predict articulation affordance~\cite{Mittal2021, Schiavi2022, Eisner2022}. The benefit of this approach is in estimating how to interact with the object, without the robot needing to physically touch it. However, this introduces additional limitations such as computational cost, poor generalization to previously unseen categories of objects, and an inability to predict articulation for objects without obvious visual cues of articulation, such as \figref{fig:motivation}. These works also lack the quality of \emph{interactive perception}~\cite{bohg_interactive_2017} present in previous motion observation methods.

In this work, partly inspired by Lips and wyffels~\cite{Lips2023}, we seek to revisit the use of proprioceptive sensing in manipulating articulated objects, while exploiting the latest deep learning advances. We propose a system that uses a neural network to predict object affordance as an initial guess and then updates the model estimate from proprioceptive sensing while interacting with the object. This is analogous to how a human might initially try to pull open a drawer only to realize during interaction that it is a revolute door.

We use a network trained in simulation using the Partnet-Mobility dataset~\cite{Mo2018} to predict articulation affordance from an initial point cloud. Then, at the grasp point specified by a user, the robot will interact with the articulated object and use proprioceptive sensing to estimate the articulation parameters while opening. We formulate this estimation problem as a factor graph and estimate screw parameters which are then passed to our symbolic math model for motion generation. In summary, the contributions of this paper are:

\begin{itemize}
    \item Online estimation of articulation parameters using vision and proprioceptive sensing in a factor graph framework.
    \item Full system integration for shared autonomy between a human user and the robot for opening articulated objects.
    \item Validation of our system with extensive real-world experimentation, opening several articulated objects with the estimation running in a closed loop.
\end{itemize}

\section{Related work}
In this section, we briefly summarize related works on articulation estimation. First, we cover interactive perception~\cite{bohg_interactive_2017} methods which have a long history of use with articulated objects. Then, we briefly cover the most relevant, recent deep learning methods for vision-based articulation prediction.

\subsection{Interactive Perception Methods}

Few works use solely proprioception for estimating articulation due to challenges in identifying a grasp point. Jain~\etal~\cite{Jain2010} simplified the problem by assuming a prior known grasp pose and initial opening force vector which allowed them to open several everyday objects. They demonstrated that once a robot is physically interacting with an articulated object and is given a good initial opening direction, proprioception alone can be sufficient to open the objects.

More commonly, proprioception is fused with vision. Sturm~\etal~\cite{Sturm2011} introduced a probabilistic framework for estimating articulation. Their method explicitly classifies the type of joint by maximizing the likelihood from an observed trajectory. They tested their method using fiducial markers, depth images, and kinematic information. Later, other work proposed a method based on bundle adjustment of visual features~\cite{katz_interactive_2014}. Martin-Martin~\etal~\cite{MartinMartin2022} introduced a framework that can estimate online from vision and tactile sensing. Like previous work, they track the motion of visual features while the robot is interacting with the object. This is fused with force/torque sensing, haptic sensing from a soft robotic hand, and end effector pose measurements.

Heppert~\etal~\cite{Heppert2022} proposed a neural network to track the motion of the parts from vision. The tracked poses are connected by a factor graph to estimate the joint parameters. In their experiments, they estimated an unknown articulation; however, their controller used a prescribed motion to open the object, giving sufficient information to the estimator. In our work, we also use a factor graph that connects part poses to a joint screw model. However, we use predictions from a neural network to give the robot an initial estimate of the articulation which is then updated online from kinematic sensing. Our use of both interactive perception and learning-based predictions allows us to perform closed-loop control and estimation while opening unknown articulated objects.

\subsection{Learning-Based Methods}
Recently, several works have focused on using only visual information with deep learning to provide a prediction on articulation without needing to interact with the object. Many of these works are trained using simulated datasets such as the PartNet-Mobility dataset~\cite{Mo2018} which contains examples of common articulated objects. Articulation-aware Normalized Coordinate Space Hierarchy (ANCSH)~\cite{Li2019} takes a point cloud as input and predicts the part poses as well at the joint parameters. However, this method requires prior knowledge of the object category (e.g. oven or refrigerator).

Most works focus on learning category-free articulation affordances~\cite{mo_where2act_2021, xu_umpnet_2022, Eisner2022} which describe what a user can do with an object. In the case of articulated objects, this is typically parameterized as a normalized vector which describes the motion of a point on the articulated part of an object. Bahl~\etal~\cite{Bahl2023} used a neural network to learn grasp pose and opening trajectory together from prior human demonstrations.
They demonstrated their method with an impressive range of real robots and articulated objects.

All of these works have the same limitations which makes their use on real robots challenging. They use only visual information and have large computational requirements which prevents online estimation. When they are used with real data, they typically take a single ``snapshot'' of the object and make a single inference. However, due to the reliance on recognizing visual similarity in objects compared to past training experiences, these methods exhibit poor performance on an object like in \figref{fig:motivation} which has no visual indicators as to how it opens. If the predictions are wrong, then these methods are reliant on highly compliant controllers to account for the error due to a lack of online estimation. 

\subsection{Systems}
In our work, we provide not only an estimation method but also a full system for opening articulated objects with shared autonomy. Therefore, we also mention some related systems work. Mittal~\etal~\cite{Mittal2021} introduced a system for whole-body mobile manipulation. They used the category-level object pose prediction network from~\cite{Li2019}. This meant their method needed prior information about the category of object with which the robot interacted. Also, in their method, they make a single prediction before interaction and then rely on controller robustness to account for mistaken predictions.

A closed-loop learning estimation method was proposed by Schiavi~\etal~\cite{Schiavi2022}. This method estimates articulation affordance from vision at multiple time steps during the interaction. A sampling-based controller solves for the optimal opening trajectory. When opening the object becomes stagnant due to torque limits, the robot releases the object and moves to a configuration to view the full object again, then makes a new vision-based estimate of the articulation.

These systems rely heavily on robust and compliant controllers to account for all errors in articulation estimation. In contrast, our work updates the estimation of the articulation model seamlessly during interaction.

\section{Screw Theory Background}
% https://ethz.ch/content/dam/ethz/special-interest/mavt/robotics-n-intelligent-systems/multiscaleroboticlab-dam/documents/trm/HS2019/Slides/03_2019-10-14_ScrewTheory.pdf
% https://ethz.ch/content/dam/ethz/special-interest/mavt/robotics-n-intelligent-systems/multiscaleroboticlab-dam/documents/trm/HS2019/TRM19.pdf
Screw theory is the geometric interpretation of twists that can be used to represent any rigid body motion (Chasles theorem)~\cite{Murray1994}. Screw motions are parameterized by the twist $\xi = ( \linvel, \rotvel ), \text{where} ~ \linvel, \rotvel \in \R^3$. The variable $\linvel$ represents the linear motion and $\rotvel$ the rotation.
We can convert this to a tangent space to $\SEthree$ using $\hat{\xi}$ as
\begin{equation}
	\label{eq:xi_hat}
	\hat{\xi} = \begin{bmatrix}
	\hat{\rotvel} & \linvel\\
	0 & 0
\end{bmatrix} \in \mathfrak{so}(3),
\end{equation}
where the hat operator $\hat{(\cdot)}$ is defined as:
\begin{equation}
	\label{eq:hat}
	\hat{\boldsymbol{\omega}} = \begin{bmatrix}
	0 & -\omega_z & \omega_y\\
	\omega_z & 0 & -\omega_x\\
	-\omega_y & -\omega_x & 0
\end{bmatrix}.
\end{equation}

This can be converted to the homogeneous transformation $\mathbf{T}^{twist}(\hat{\xi}, \theta) \in \SEthree$ using the exponential map Exp: 
\begin{align}
    \label{eq:expmap}
    \mathbf{T}^{twist}(\hat{\xi}, \theta) = \text{Exp}(\xihat\theta),
\end{align}
where $\theta \in \R$ is the articulation configuration. 

Now if we define a fixed world frame $\World$, the pose of the moving part of an articulated object $\mathbf{T}_\mathtt{WA} \in \SEthree$ is related to the other, non-moving part $\mathbf{T}_\mathtt{WB} \in \SEthree$ by
\begin{align}
\label{eq:transform}
    \mathbf{T}_\mathtt{WA} = \mathbf{T}_\mathtt{WB}\mathbf{T}^{twist}(\hat{\xi}, \theta).
\end{align}

\section{Problem Statement}
The goal of this work is to estimate online the Maximum-A-Posteriori (MAP) state of a single joint from visual and proprioceptive sensing. We define the state $\State(t)$ at time $t$ as
\begin{equation}
\State(t) \triangleq \left[\xi,\theta(t) \right] \in \Real^{7},
\end{equation}
where $\xi$ are the screw parameters which are constant for all time and $\theta(t)$ is the angle of articulation at time $t$. We assume the object is composed of only two parts connected by a single joint. This encompasses the vast majority of articulated objects and therefore is a reasonable simplification. We define the pose of each part in the world frame $\World$ as $\partA_{\World\mathtt{A}}, \partB_{\World\mathtt{B}} \in \SEthree$ where $\partB_{\World\mathtt{B}}$ is the bottom part that is static and $\partA_{\World\mathtt{A}}$ is the articulated part which the robot grasps. 
We estimate $\mathsf{K}$ poses at time indices $k$; so the set of all estimated states and articulated part poses can be written: $\calX = \{\State_k, \partA_{k}, \partB_{k}\}_{k \in \mathsf{K}}$, dropping reference frames for clarity.

We use $\mathsf{P}$ point clouds at times $p$ which are each associated with a prediction on $\xi$. 
Without loss of generality, we set $\mathsf{P} = 1$ with one visual measurement at the beginning, although in future work we could add multiple predictions. 
We use $\mathsf{K}$ kinematic measurements at times $k$ which are each associated with a pose estimate of the grasp point. The times $k$ are only selected while the robot is in contact with the object and after the articulated part has been moved a certain distance $d$ to avoid taking too many measurements. The set of all measurements are then grouped as $\calZ = \{\calP_{p}, \calK_{k}\}_{p \in \mathsf{P}, k \in \mathsf{K}}$ where $\calP$ are the point clouds and $\calK$ the pose measurements from kinematics. 

\begin{figure}[t]
	\centering
	\includegraphics[width=\columnwidth]{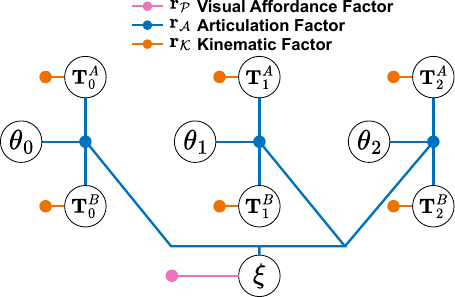}
	\caption{The factor graph shows the variables we are estimating: $\partA(t), \partB(t), \theta(t)$ and $\xi$ which exists at only one time step in the factor graph. We show three time steps including the initial visual affordance factor which provides a prior estimate on $\xi$ as a unary factor.}
	\label{fig:fg}
\end{figure}

\begin{figure*}[t!]
	\centering
	\includegraphics[width=0.8\linewidth]{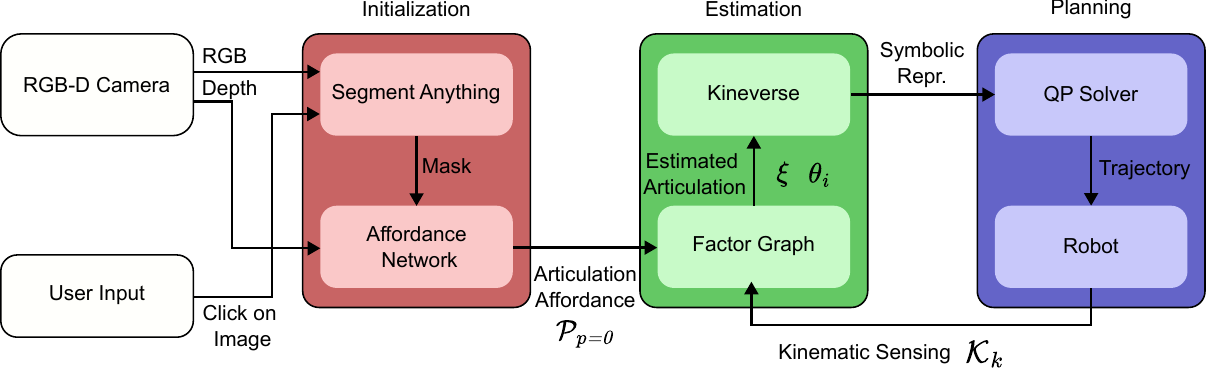}
	\caption{Full system with information flow. An RGB-D camera provides RGB images which are segmented with the click prompt from a human user. This generates a mask on the articulated part which, with depth information from the camera predicts initial articulation parameters. This is provided to the factor graph which also uses kinematic measurements of the end effector to estimate the object articulation. The estimated articulation updates the symbolic math representation of both robot and object which is then formulated as a quadratic programming (QP) problem to solve for robot trajectory.}
	\label{fig:pipeline}
\end{figure*}

\section{Factor Graph Formulation}
We maximize the likelihood  of the measurements $\calZ$, given the history of states $\calX$:
\begin{equation}
	\calX^* = \argmax_{\calX} p(\calX|\calZ) \propto p(\State_0)p(\calZ|\calX),
	\label{eq:posterior}
\end{equation}
where $\calX^*$ is our MAP estimate of the joint.

We assume the measurements are conditionally independent and corrupted by zero-mean Gaussian noise. Therefore, \eqref{eq:posterior} can be expressed as the following least squares minimization:
\begin{align}
\begin{split}
	\calX^{*} =& \argmin_{\calX} \|\mathbf{r}_\textup{0}\|^2_{\Sigma_\textup{0}} + \\
	               &\sum_{p \in \mathsf{P}} \|\mathbf{r}_{\calP_{p}}\|^2_{\Sigma_{\calP}}
	               +  \sum_{k \in \mathsf{K}} \Big( \|\mathbf{r}_{\calA_{k}} \|^2_{\Sigma_{\calK}} + \|\mathbf{r}_{\calK_{k}} \|^2_{\Sigma_{\calK}} \Big),
	\label{eq:cost-function}
\end{split}
\end{align}
where each term is the residual $\mathbf{r}$ associated with a measurement type, weighted by the inverse of its covariance matrix. A factor graph can be used to graphically represent \eqref{eq:cost-function} as shown in Fig~\ref{fig:fg} where large white circles represent the variables we would like to estimate and the smaller colored circles represent the residuals as factors. The implementation of the factors is detailed in Sec.~\ref{sec:estimation}.

\section{Proposed Method}
This section describes the full system as shown in Fig~\ref{fig:pipeline}.

\subsection{Initialization}
For the initialization module, we use the latest advances in deep learning for articulated objects and introduce a system of shared autonomy. First, a user is presented with a video feed of the object and clicks on the desired grasp point. With this query point, we use Kirillov \etal's Segment Anything (SAM)~\cite{Kirillov2023} to segment a mask\footnote{As an additional, minor contribution, we open source the ROS wrapper for SAM which was used for this work~\url{https://github.com/robot-learning-freiburg/ros_sam}} of the non-static part.

The image mask and associated point cloud are then passed to the network which predicts the articulation affordance for each masked point.
This affordance is parameterized as a point cloud with the predicted, normalized acceleration of each point given a force opening the object.
This idea was first introduced by Zeng~\etal~\cite{Zeng2021} where the concept was described as motion residual \emph{flow} and later improved by Eisner \etal with Flowbot3D~\cite{Eisner2022}.
Our network is identical to Flowbot3D but, importantly, we change the prediction of flow to be in the camera frame, not the global frame.
We found this necessary to enable the method to work with real objects of arbitrary pose, otherwise, the predicted flows were always directed away from the camera's optical center.
Therefore, the output of the initialization step is the point cloud affordance $\calP_{p=0}$, and the 3D point associated with the user's click which will be used as the first planning goal for the robot.

\subsection{Factor Graph Estimation}
\label{sec:estimation}
For estimating the screw parameters, we use the factor graph in Fig.~\ref{fig:fg}. This is similar to~\cite{Heppert2022}, however, instead of visually tracking the different parts we use kinematic sensing which enables online estimation. We also introduce a new articulation affordance factor to integrate the visual-based prediction from a neural network. Our method makes no assumption about what type of articulation is being estimated, e.g. revolute or prismatic.

\subsubsection{Affordance Factor}
To incorporate the predicted affordances we introduce an articulation affordance factor. First, we fit a plane to the affordance point cloud $\calP$. We then create a new point cloud by adding the affordance to each point in $\calP$, scaled by a small increment. This results in a second point cloud $\calP^{+}$ which has been slightly shifted as though the object were opened. We fit a second plane to $\calP^{+}$.

\begin{figure}[t]
	\centering
	\includegraphics[width=\columnwidth]{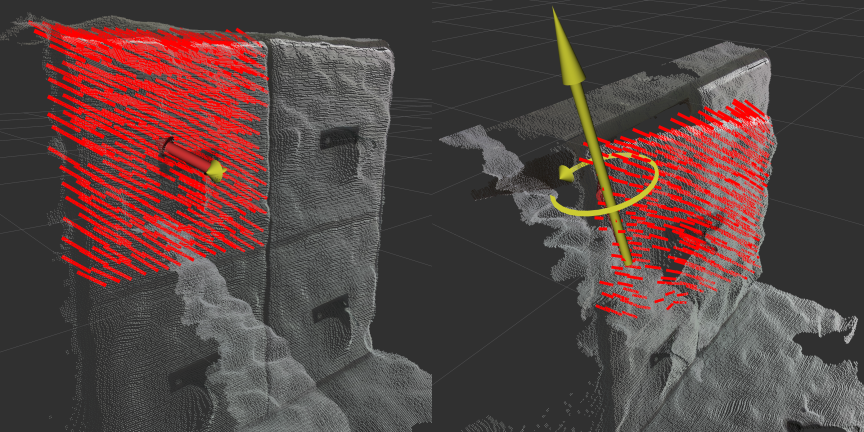}
	\caption{Example affordance predictions from the neural network: prismatic left and revolute right. The small red lines are the output of the network, predicting articulation flow on the segmented points. The large red and yellow arrows indicated the resulting joint prediction.}
	\label{fig:network-output}
\end{figure}

From the two planes, we find the intersecting line from the cross product of the normals. This defines the axis of rotation of a revolute joint. If the cross product is zero, or very small, then we assume the joint is prismatic as shown in Fig.~\ref{fig:network-output}. A similar approach was done by Zeng~\etal~\cite{Zeng2021} and results in a prediction $\tilde{\xi}$ on the articulation which can be used in \eqref{eq:cost-function} as the affordance factor:
\begin{align}
    \mathbf{r}_{\calP_{p}} = \xi - \tilde{\xi}.
\end{align}

In future work we intend to train the network to predict its own uncertainty similar to~\cite{Russell2021}, however, at the moment we select a fixed value ($\sigma_{\calP} = 1e^{-3}$).

\subsubsection{Articulation Factor}
From inspection of (\ref{eq:transform}), we can see that all variables in the factor graph are related. The articulation residual can then be computed in \eqref{eq:cost-function} as:
\begin{align}
    \mathbf{r}_{\calK_{A}} = \mathbf{T}^{twist}(\xihat, \theta_k) \boxminus {\partB_k}^{-1} \partA_k,
\end{align}
where $\boxminus$ is a pose differencing over the manifold using the logarithm map:
\begin{align}
    \partA \boxminus \partB = \text{Log}({\partB_k}^{-1} \partA_k) \in \mathfrak{so}(3).
\end{align}

\subsubsection{Kinematic Factor}
Kinematic measurements are added to the graph as unary pose factors on $\partA$ and $\partB$. We assume $\partB$ doesn't move and reuse the initial grasp pose. The residual $\mathbf{r}_{\calK_{k}}$ is the default $\SEthree$ factor in GTSAM~\cite{gtsam}.

\subsection{Online Motion Generation}
This subsection covers the computation of the desired robot configurations~$\vec{q}\in\Real^7$, given the latest estimate of the articulation~$\xi$.
We model the forward kinematics of the robot end-effector as~$\mat{T}_{\World\EndEff}(\vec{q})$, and the forward kinematics of our the articulated object as
\begin{align}
\mat{T}_{\World\Articulation}(\theta, \xi) &= \T_{\World\Grasp} \cdot \mat{T}^{twist}(\theta, \xi),
\end{align}
with~$\T^{twist}$ computed using ~\eqref{eq:expmap} with the latest estimated $\xi$ and a goal $\theta$. The pose $\T_{\World\Grasp}$ is a static transformation composed of the user defined grasp position~$\vec{p}_{\World\Grasp}$ and a predetermined grasp orientation~$\vec{R}_{\World\Grasp}$.

Once the robot grasps the object handle, we set $\theta_0 = 0$, which leads to $\mat{T}^{twist}(0, \xi_0) = \I$.
We then progressively increment the desired articulation configuration $\theta_{t + 1} = \theta_t + gv \Delta t$, with $gv$ being a constant speed for opening/closing the articulation, up to the articulation limit after which we invert the sign of~$gv$.
For each $\theta_t$, and given an estimate of $\xi$, we solve the inverse kinematics (IK) problem, subject to the condition~$\mat{T}_{\World\EndEff}(\vec{q}_{t+1}) = \mat{T}_{\World\Articulation}(\theta_{t+1}, \xi)$.
More specifically, we define the IK problem as a non-linear optimization problem where we encode the following task space constraints
\begin{align}
\begin{split}
    \norm{\mat{p}_{\World\EndEff}(\vec{q}) - \mat{p}_{\World\Articulation}(\theta_t, \xi))}_F^2 &= 0\\
    \norm{\mat{R}_{\World\EndEff}(\vec{q}) - \mat{R}_{\World\Articulation}(\theta_t, \xi))}_F^2 &= 0
\end{split}
\label{eq:pose_constraint}
\end{align}
where~$\norm{\cdot}_F$ denotes a Frobenius norm.

We exploit the differentiability of the constraints in~\eqref{eq:pose_constraint} w.r.t.~to \vec{q}, to linearize the problem, and solve it sequentially until constraint satisfaction as a quadratic program (QP):
\begin{equation}
    \begin{aligned}
        \argmin_{\mathbf{x}}\,\frac{1}{2}\mathbf{x}^T\mathbf{C}\mathbf{x} \quad
        \textrm{s.t.} \quad & \mathbf{lb}   \leq \mathbf{x}  \leq \mathbf{ub} \\
                        & \mathbf{lb}_A \leq \mathbf{Ax} \leq \mathbf{ub}_A,
    \end{aligned}
    \label{eq:qp}
\end{equation}
where $\vec{x} = \langle \vec{\fd{q}}, \vec{s}\rangle$ is a vector of joint velocities and slack variables $\vec{s}$, and $\mat{A}$ is the Jacobian of the task constraints and association with the slack variables.
\eqref{eq:qp} also encodes bounds on robot joint positions and velocities.

We use the Kineverse articulation model framework~\cite{Rofer2022} for representing both the robot and the articulated object forward kinematics and constraints, computing the Jacobians, as well as encoding and solving the problem in~\eqref{eq:qp}.
Kineverse uses the CasADi symbolic math backend~\cite{Andersson2019}, enabling effortless computation of gradients for arbitrary expressions, such as the articulation model.

Finally, we command the resulting joint positions $\vec{q}_{t+1}$ to the robot in compliant mode.
Therefore, if the articulation estimation~$\xi$ is inaccurate, the robot can comply with the physical articulation, leading to an end-effector pose that is different from~$\T_{\World\Grasp}(\theta_{t+1},\xi)$.
The end-effector pose~$\mat{T}_{\World\EndEff, t+1}$ is added to the graph as a measurement on $\partA$.

\section{Experiments}

\subsection{Implementation details}
For experiments, we used the compliant KUKA LBR iiwa robot.
We used an Intel Realsense D435 camera and Robotiq 140 two finger gripper. We implemented the factor graph using the GTSAM library~\cite{gtsam}. 

\subsection{Hand Guiding Experiments}
\label{sec:hand-guiding}
In these experiments, we investigated the accuracy of our factor graph estimation module. We compared our method against Heppert \etal\cite{Sturm2011} which also uses factor graphs to estimate a screw parameterization. The authors kindly granted us access to their code for direct comparison.

\begin{figure}[t]
	\centering
	\includegraphics[width=\columnwidth]{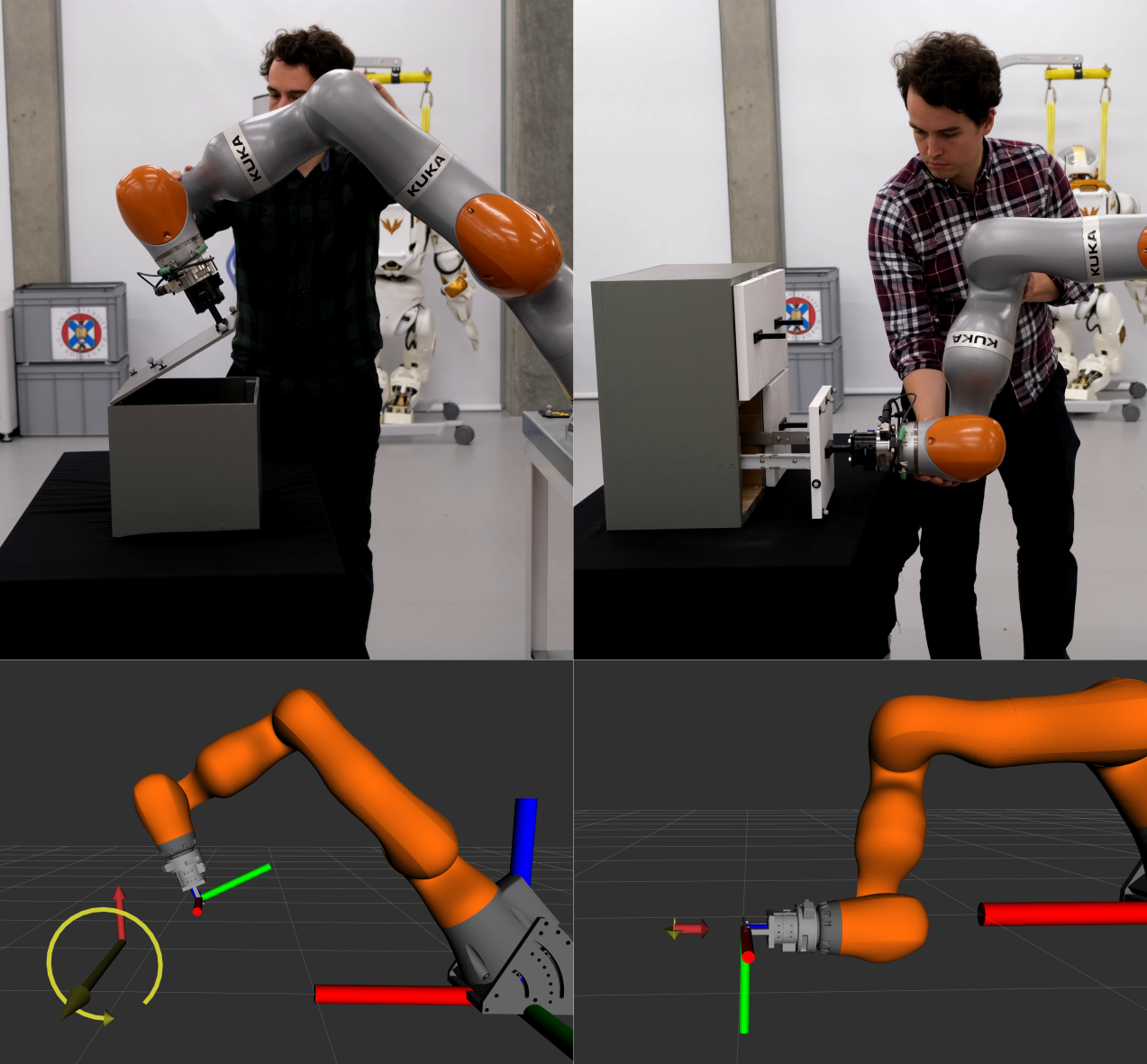}
	\caption{Top: hand guiding experiments for revolute (left) and prismatic (right) joints. Bottom: the resulting estimated articulation. Yellow arrows show $\rotvel$ while red arrows show $\linvel$. The large axis is the base frame of the robot which is used for $\World$ while the small axis is the estimated pose $\partA$.}
	\label{fig:hand-guiding-experiment}
\end{figure}

We physically attached the robot's end-effector to the box lid and hand-guided the robot motion, in gravity compensation mode, to open and close the box.
For this experiment, we recorded both the robot joint positions, measured by the encoders, and the respective box lid poses, tracked with Vicon motion capture, as shown in Fig.~\ref{fig:hand-guiding-experiment}. 

Similar to~\cite{Heppert2022} we use the tangent similarity metric:
\begin{align}
    J(\linvel_{gt}, \linvel_{est}) = \frac{1}{\theta_{max} - \theta_{min}} \int_{\theta_{min}}^{\theta_{max}} \frac{\linvel_{gt}}{\norm{\linvel_{gt}}} \cdot \frac{\linvel_{est}}{\norm{\linvel_{est}}},
\end{align}
where $\linvel_{gt}$ is the local linear velocity of the grasp point measured from Vicon and $\linvel_{est}$ is the estimated local velocity from the articulation model. We can compute $\linvel_{est}$ from $\xi$ using the equation: $\linvel_{est} = \linvel + \rotvel \times \vec{c}$ where $\vec{c}$ is the contact point from kinematics. Since $\linvel_{gt}$ and $\linvel_{est}$ are normalized, they represent the direction of motion; therefore, their tangent similarity will be 1 when identical and 0 when perpendicular.

\begin{figure}[t]
	\centering
    \includegraphics[width=\columnwidth]{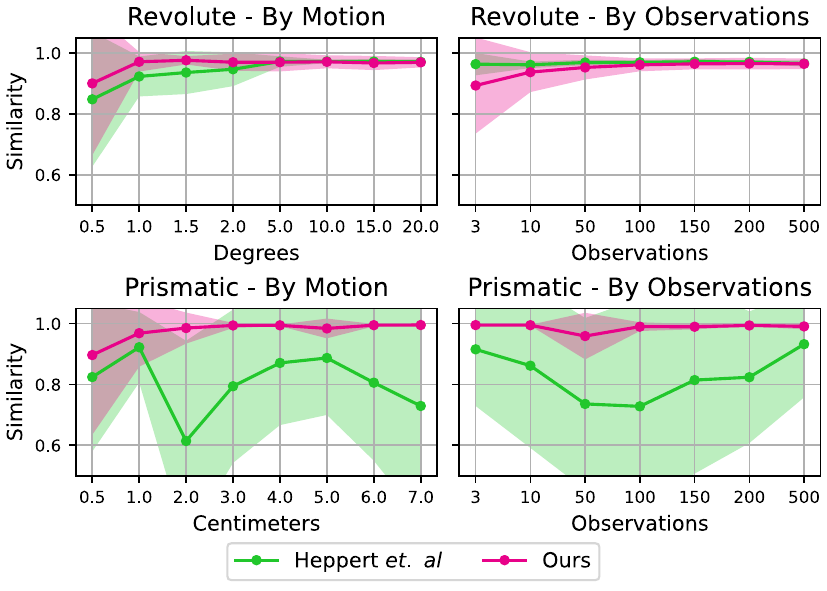}
	\caption{Tangent similarity for hand guiding experiments. The solid line shows average error while the shaded region shows standard deviation.}
	\label{fig:hand-guiding-results}
\end{figure}

We recorded two hand guiding experiments, one for a revolute joint and one for a prismatic. First, we performed optimization over fixed increments, for example, optimizing over every \SI{1}{\degree} of rotation or \SI{1}{\centi\meter} of translation. Next, we tested using fixed numbers of measurements equally spaced over the entire configuration range with full results shown in Fig.~\ref{fig:hand-guiding-results}. In the factor graph, we make no distinction between prismatic or revolute. When estimating prismatic joints, $\rotvel$ tends towards very small values. At the output, if $\norm{\rotvel} < 0.01$ we set $\rotvel = 0_{3x1}$ and normalize $\linvel$.

We achieve good accuracy within a short window of measurements: after only \SI{0.5}{\degree} of rotation our estimator has an average tangent similarly of 0.90, after \SI{1.0}{\degree} this improves to 0.97.
This enables online re-estimation in cases where the neural network prediction is wrong since the door will only need to be moved a small amount for the correct articulation to be estimated.
Additionally, we show that for equally spaced measurements throughout the configuration range, as few as 3 measurements can be sufficient to accurately estimate the joint.
In comparison with Heppert \etal, both methods have similar performance for revolute joints while our method is better at distinguishing prismatic joints. This is likely because we check for the prismatic case whereas their method tended to confuse prismatic joints with very large revolute.

\subsection{Full System Experiments}
In these experiments, we tested the full pipeline, using vision and proprioceptive sensing together. The experiment protocol was as follows: the human user views the robot's camera feed which is looking at the same cabinet as in Fig.~\ref{fig:motivation} and clicks on the image where to gasp. The robot then moves to the grasp goal and closes the gripper. Next, the robot moves using the learned articulation prediction from $\theta = 0$ to a specified upper bound. Estimation runs online, using kinematic sensing to update the model which is fed back to the controller in a closed loop. Eventually, the estimate converges to the correct estimate of the joint and the controller continues to open and close the door. We used a distance limit of $d=\SI{2}{\milli\meter}$ or $d=\SI{0.5}{\degree}$ to trigger adding a kinematic measurement to the factor graph. We performed a new optimization after every 20 new measurements.

\begin{figure}[t]
	\centering
	\includegraphics[width=\columnwidth]{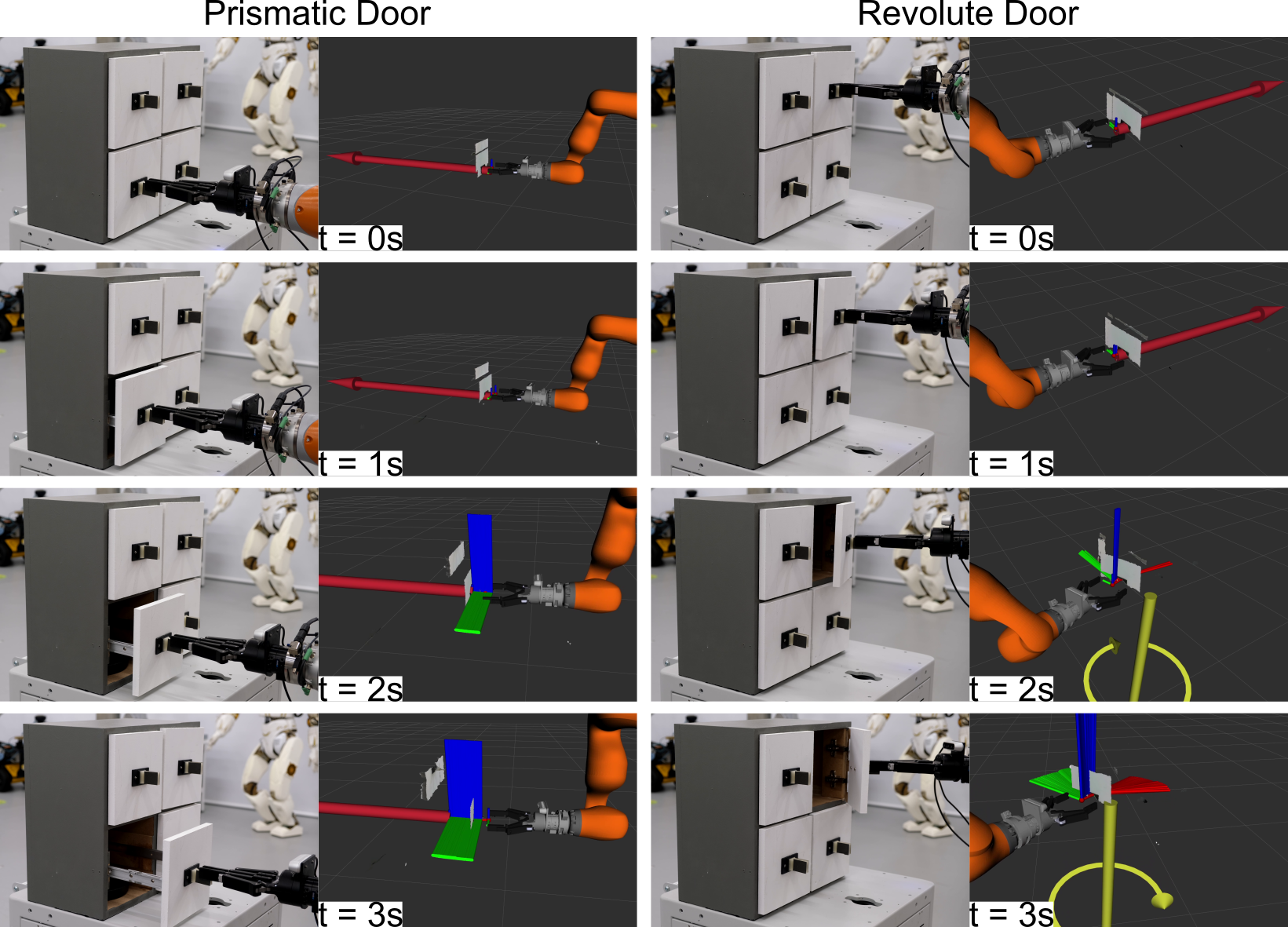}
	\caption{Real-robot experiments: the red arrows indicate the estimated $\linvel$, yellow arrows show $\omega$ and are centered on the point $q$ which lies on the axis of articulation. The axes show the measurements of $\partA$. The left column shows the robot opening the prismatic drawer using an initial neural network prediction of the prismatic joint. The right column shows the robot opening the top right revolute door with an initial neural network prediction of prismatic.}
	\label{fig:full-experiment}
\end{figure}

Two online estimation experiments are shown in Fig.~\ref{fig:full-experiment}. On the left, the robot initially received a correct prediction from the network that the joint was prismatic.
Then, as the robot opened the drawer, the estimate of the joint was refined.
On the right, the network incorrectly predicted prismatic and so the robot pulled backward on the handle. 
Because of compliance in the robot joints, the door opened by a few degrees at \SI{1}{\second}.
Opening the door by this small amount, rotated the end effector and allowed the correct articulation to be estimated from kinematics.
By \SI{3}{\second} the joint estimate correctly converged and the robot was able to fully open the door.
We repeated the full pipeline experiment 20 times on different doors and successfully opened the doors 16 times.

\section{Discussion}
Our method is able to accurately estimate articulation parameters so long as the door is able to move by a small amount, for example in the right column of Fig.~\ref{fig:full-experiment}.
While this can correct for a poor neural network prediction, one limitation to our work is that certain incorrect predictions can lead to no motion of the robot in compliant mode, such as predicted articulations with an orthogonal direction of motion to the true articulation.
For example, the robot was never able to open the bottom right door in Fig.\ref{fig:motivation}, which slides open to the right.
This is because the network always predicted either a prismatic articulation, as a drawer, or a revolute articulation.
As a result, the robot arm was unable to move and gain information when trying to open the door
To account for this, in future work we will add an ``exploration'' module to the system.
This could randomly apply forces in different directions until the arm is able to move and is similar to how humans discover articulations that are visually ambiguous.

Of the 20 full system trials attempted, 4 failed due to slipping of the gripper. Because we use kinematic measurements, we cannot differentiate between slipping and correct movement opening a door. In future work, we will investigate adding force/torque sensing and use a suction gripper.

\section{Conclusion}
In this work, we present a complete system for opening of articulated objects using shared autonomy for which visual cues alone are insufficient to correctly estimate the articulation parameters.
Our factor graph-based method incorporates both vision-based affordance predictions from a neural network with kinematic sensing in an analytical model for online estimation.
We validated the system on a real robot and opened several articulated objects, using the online estimate in a closed loop.
We succeeded in 16/20 consecutive trials and properly estimated the articulation, despite of the original affordance prediction being incorrect for 13/16 of those trials.
Nevertheless, in those cases, the robot was able to succeed due to the online re-estimation of the articulation through the compliant interaction.
In future work, we will incorporate an exploration module to enable the discovery of articulations that might be undetectable from the network predictions.

\bibliographystyle{IEEEtran}
\bibliography{ref}

\end{document}